\documentclass[10pt,twocolumn,letterpaper]{article}

\usepackage{iccv}
\usepackage{times}
\usepackage{epsfig}
\usepackage{graphicx}
\usepackage{amsmath}
\usepackage{amssymb}
\usepackage{hyperref}

\iccvfinalcopy 

\def\iccvPaperID{14} 

\ificcvfinal\fi

\begin{document}

\title{Using Segmentation Masks in the ICCV 2019 Learning to Drive Challenge}

\author{Antonia Lovjer\textsuperscript{*}\\
Columbia University \\
{\tt\small al3273@columbia.edu}
\and
Minsu Yeom\textsuperscript{*}\\
Columbia University \\
{\tt\small my2582@columbia.edu}
\and
Benedikt D. Schifferer \textsuperscript{1}\\
Columbia University \\
{\tt\small bds2141@columbia.edu}
\and
Iddo Drori \textsuperscript{2}\\
Columbia University \\
{\tt\small idrori@cs.columbia.edu}
}
\maketitle
\ificcvfinal\thispagestyle{empty}\fi

\makeatletter{\renewcommand*{\@makefnmark}{}
\footnotetext{* Columbia University Deep Learning Course Participants}\makeatother}

\makeatletter{\renewcommand*{\@makefnmark}{}
\footnotetext{1 Columbia University Deep Learning Course Assistant}\makeatother}

\makeatletter{\renewcommand*{\@makefnmark}{}
\footnotetext{2 Columbia University Deep Learning Course Instructor}\makeatother}

\begin{abstract}
In this work we predict vehicle speed and steering angle given camera image frames. Our key contribution is using an external pre-trained neural network for segmentation. We augment the raw images with their segmentation masks and mirror images. We ensemble three diverse neural network models (i) a CNN using a single image and its segmentation mask, (ii) a stacked CNN taking as input a sequence of images and segmentation masks, and (iii) a bidirectional GRU, extracting image features using a pre-trained ResNet34, DenseNet121 and our own CNN single image model. We achieve the second best performance for MSE angle and second best performance overall, to win 2nd place in the ICCV Learning to Drive challenge. We make our models and code publicly available \cite{lovjer2019learningtodrive}.
\end{abstract}


\section{Introduction}
\begin{figure}
    \includegraphics[width=1\linewidth]{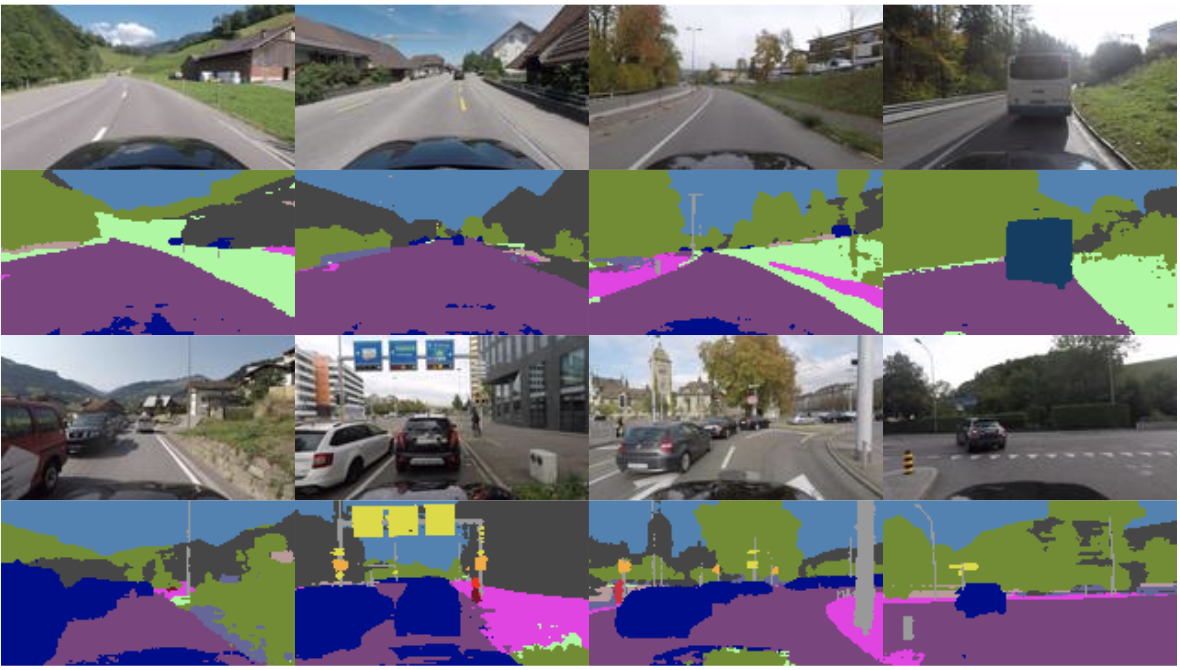}
    \caption{Sample of training images. The raw road images were provided in the ICCV competition data set, and the segmentation masks were computed using a pre-trained model from NVIDIA which was trained on the Cityscapes dataset \cite{semantic_cvpr19, Cordts_2016_CVPR}}
    \label{fig:training_image}
\end{figure}

\begin{figure}
    \includegraphics[width=1\linewidth]{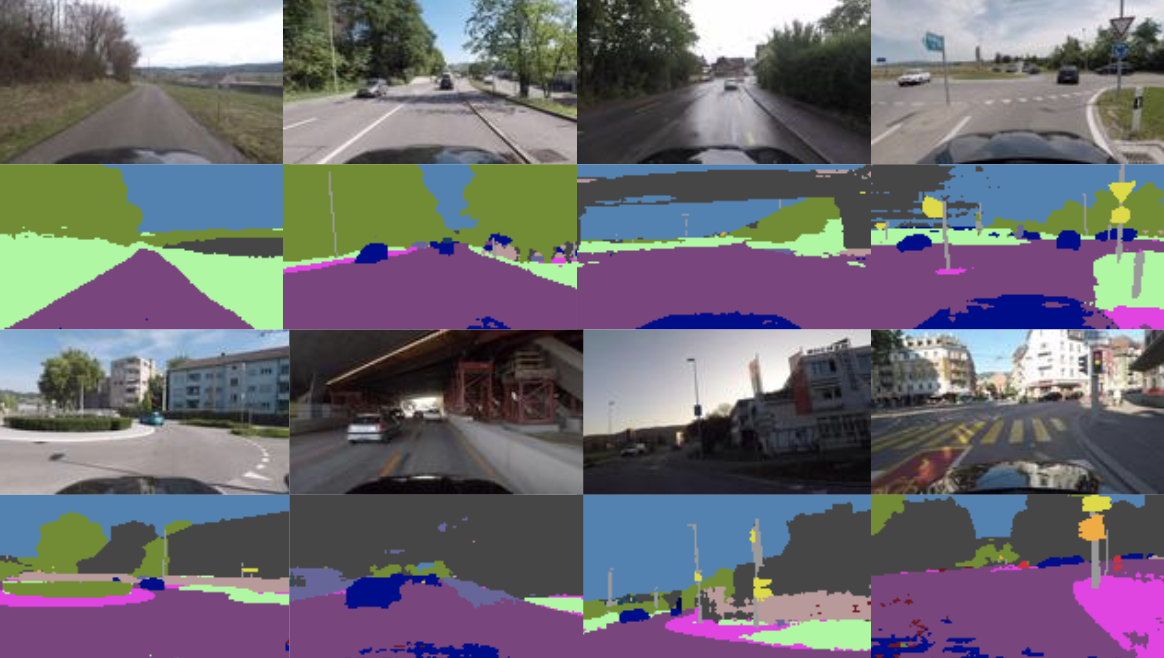}
    \caption{Sample of test images. The raw road images were provided in the ICCV competition data set, and the segmentation masks were computed using a pre-trained model from NVIDIA which was trained on the Cityscapes dataset \cite{semantic_cvpr19, Cordts_2016_CVPR}}
    \label{fig:testing_image}
\end{figure}

We show that jointly predicting the future steering angle and vehicle speed may be improved through segmentation maps and data augmentation techniques. Motivated by the use of semantic segmentation models for self driving vehicles \cite{xu2017end, hou2019learning}, we concatenate the segmentation maps with the images as the input instead of using the maps as an additional learning objective. We augment the data set by applying image transformations such as mirroring, adjusting brightness, and geometric transformation. We ensemble three neural network architectures to achieve our best performance. To appreciate the dataset and task, Figure \ref{fig:training_image} shows a sample of the training images and Figure \ref{fig:testing_image} shows a sample of the test images.

\label{sec:introdcution}

\section{Related Works}
CNNs have be trained to predict steering angle given only the front camera view \cite{bojarski2016end}. As humans have a wider perceptional field than the front camera, 360-degree view datasets have been collected with additional geo-locations \cite{hecker2018end, hecker2019learning}. Neural network models have been trained end-to-end using 360-degree views which are specifically useful for navigating cities and crossing intersections. As noted, map data improves the steering angle prediction \cite{hecker2018end}. Long-term dependencies can also be captured by processing the visual input and adding steering wheel trajectories into memory networks \cite{fernando2017going}. A classical method is to extract hidden features using pre-trained CNNs and process them through a sequence model, namely a LSTM. Predicting segmentation masks can be added to the loss, which improves the overall performance \cite{xu2017end}. Instead of using segmentation masks in the loss, in this work we concatenate the segmentation maps to the input.

\section{Methods}

\begin{figure}
    \includegraphics[width=1\linewidth]{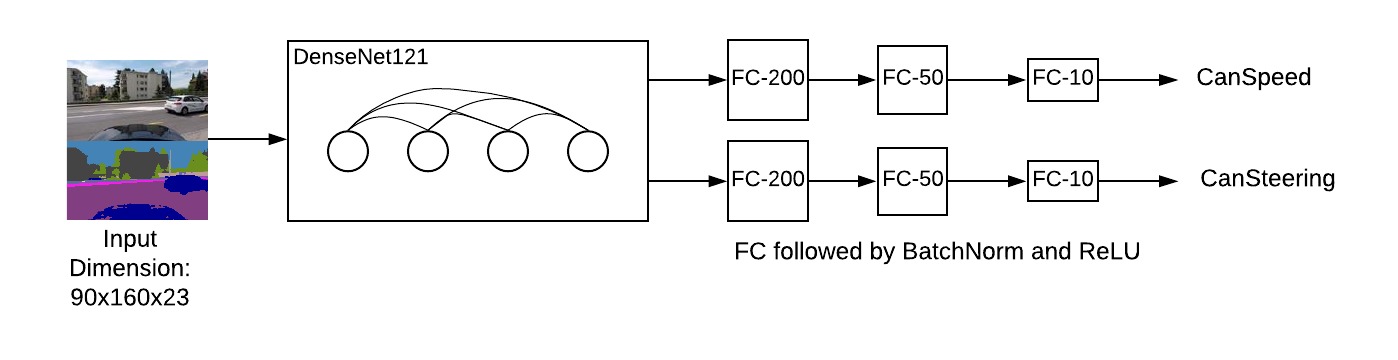}
    \caption{
    shows the architectures of $\textsc{A\_CNN\_single\_image}$. The input to the model is a single image concatenated with its hot-n encoded segmentation mask. The model input dimensions are width $\times$ height $\times$ 23 (3 RGB channels + 20 classes). The model uses a DenseNet121 architecture. The final layers are two separate feedforward passes - one for speed and for for angle. The feed forward layers, except of the output, are followed by batch normalization and ReLU.}
    \label{fig:arch_A}
\end{figure}

We implemented a number of preprocessing steps that made our model faster to train, and also more robust to variation in the input images. In particular, we used downsampling and various forms of data augmentation to alter the raw image files in the original data set before inputting it into our models. Furthermore, we will illustrate the three model architectures that we found to give the lowest MSE on speed and angle, and which were eventually combined to provide our best submission. 

One of the key innovations we focus on is using segmentation masks as inputs in addition to the raw images. Our hypothesis is that the segmentation mask cleans up a lot of unnecessary information in the image, and gives the model a clearer view of the road as well as indicators of speed. The other idea that we tried out is to avoid pretained CNNs and train them from scratch. Stacking the images to one input for a CNN trains faster than using a recurrent network. Other methods feed single images into a recurrent network, such as an LSTM. We experienced that the CNN with an LSTM cannot be trained efficiently.

These models in combination with the data preprocessing and augmentation complemented each other to provide the second best overall performance in the competition.

\label{sec:methods}

\paragraph{Data processing}
Instead of using the full dataset at 10 frames per second, we down-sampled in time 1:10 to one frame per second. We then downsampled spatially by 1:12 in each dimension from 1920$\times$1080 to 160$\times$90 for efficiency purposes. The images are also normalized with the mean and standard deviations in the training set.

Using the NVIDIA semantic segmentation model pre-trained on the Cityscapes dataset \cite{semantic_cvpr19}, we created segmentation masks for each of the images in the dataset. The original pre-trained model contained 34 classes, but as we only wanted to focus on objects that might influence the steering angle, we took only 19 classes including road, car, parking, wall, etc. \cite{Cordts_2016_CVPR}. 

\paragraph{Data augmentation}
The data underwent several types \cite{towardsdatascience} of data augmentation that were randomly administered to 80\% of the training data.

\begin{itemize}
    \item Randomly flip the images horizontally with probability 0.5 (steering angle is multiplied by -1 in order to offset the horizontal flip)
    \item Randomly change the brightness by a factor between 0.2 and 0.75 with probability 0.1
    \item Randomly shift the image left/right/up/down and adjust the angle with left/right shift with probability 0.25
\end{itemize}

\paragraph{Network architectures}
The final model consisted of an ensemble of three models: A, B, and C. Model A, $\textsc{A\_CNN\_single\_image}$, takes as input a single image and its corresponding segmentation mask. The input is passed through a DenseNet121 architecture, then into two fully connected towers, one for predicting the speed, one for the steering angle, as visualized in Figure \ref{fig:arch_A}. The towers contain three dense layers of size 200, 50 and 10. In between each dense layer we conduct batch normalization, and apply a ReLU non-linearity. The final output is a real-valued number which is the predicted speed or steering angle, which has been denormalized given the mean and standard deviation from the training set. 

\begin{figure}
    \includegraphics[width=1\linewidth]{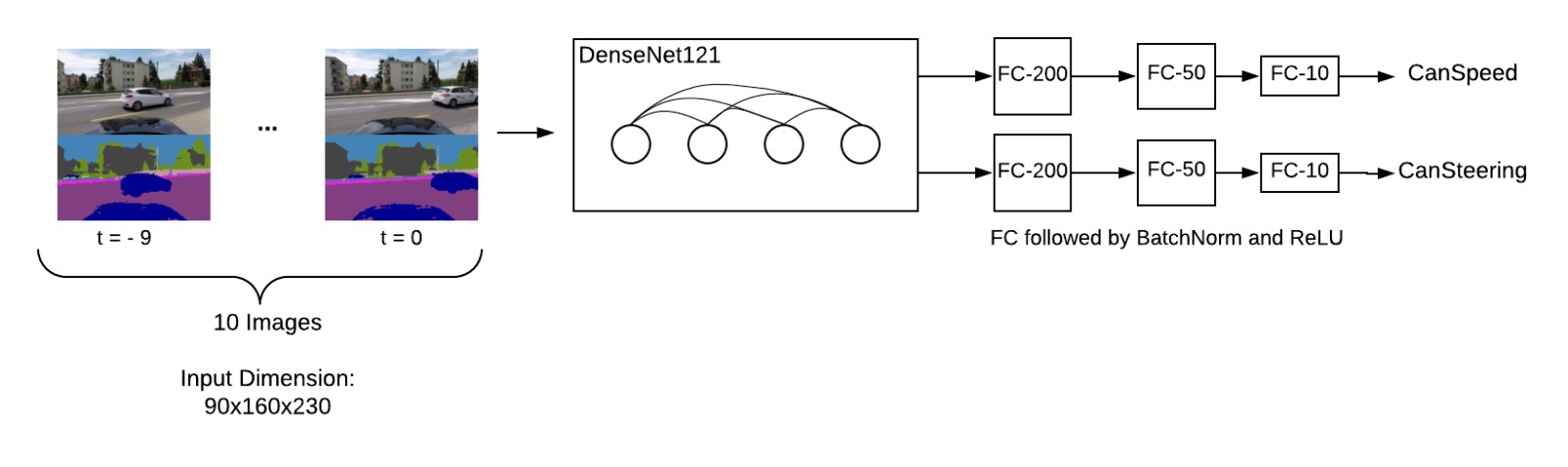}
    \caption{
    shows the architectures of $\textsc{B\_CNN\_stacked}$. Given the prediction at t=0, the input is the sequence of images from -9 to 0 - for every second one image. The input are the 10 consecutive images with their hot-n encoded segmentation masks. All images and masks are stacked to a width $\times$ height $\times$ 230 (3 RGB channels  + 20 classes)*10. The model is similar to $\textsc{A\_CNN\_single\_image}$ with DenseNet121 architecture (230 input channels) and two feed forward passes for speed and angle prediction. The feed forward layers, except of the output, are followed by batch normalization and ReLU.}
    \label{fig:arch_B}
\end{figure}

The architecture of model B is shown in Figure \ref{fig:arch_B}. $\textsc{B\_CNN\_stacked}$, takes as input a full sequence of 10 images with their corresponding segmentation masks which are concatenated together. Similar to model A, the input is passed through DenseNet121, and into the speed and steering angle regressor towers which are the same as in model A.

\begin{figure*}
    \includegraphics[width=1\linewidth]{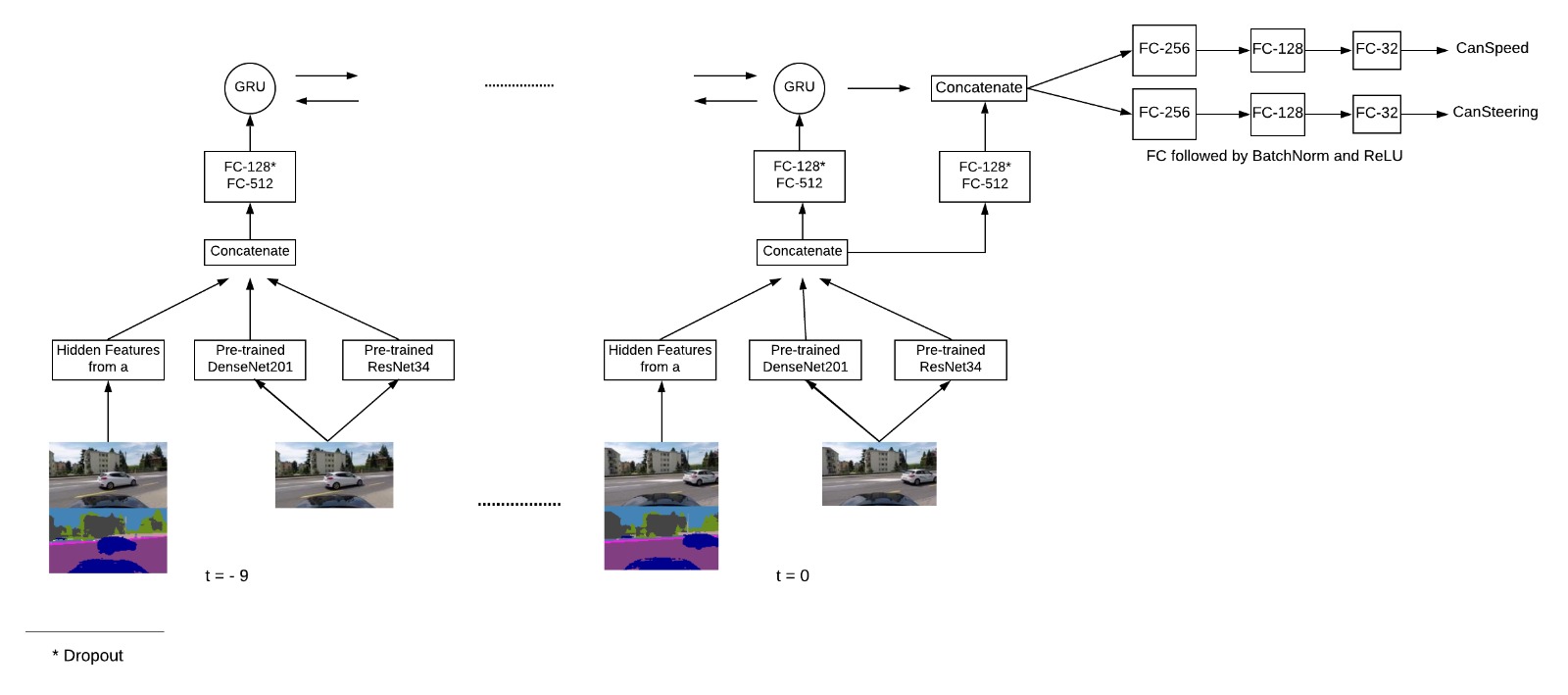}
    \caption{$\textsc{C\_Bi\_GRU}$ architecture. The input is a sequence of 10 images and its segmentation mask. For each image (+ its segmentation mask), the hidden features after the convolutional layers are extracted from pre-trained $\textsc{A\_CNN\_single\_image}$, ResNet34, and DenseNet121. The hidden features are concatenated and passed through 2 feed forward layers. The result is a sequence of features, which is fed into a 3-layered bi-directional GRU with 64 hidden features. The output of the bi-directional GRU is concatenated with the hidden features of the latest input image and its segmentation mask. Finally, the output is fed into two feed forward passes for speed and angle.}
    \label{fig:arch_C}
\end{figure*}

Figure \ref{fig:arch_C} illustrates model C, $\textsc{C\_Bi\_GRU}$. It takes as input a sequence of 10 images, and their corresponding segmentation masks, similar to model B, except that they are not concatenated together. Each image in the input sequence is passed individually through a pre-trained ResNet34 model, a pre-trained DenseNet201, and model $\textsc{A\_CNN\_single\_image}$. The models are pre-trained on ImageNet, and not on a task that directly involves vehicle images. The resulting outputs are concatenated and passed through an intermediate layer which contains two dense layers of size 512 and 128 (with dropout). The output is then passed into a bi-directional GRU cell. This occurs for every input image/mask pair in the sequence, and the output of the final GRU cell is concatenated with the input to the previous fully connected layer which also passed through a dense block. This is then fed into the two towers for the speed and steering angle prediction, where the size of the layers is 256, 128 and 32. 

\subsection{Implementation}
We used a similar method to train all the models, with slight changes in learning rate and learning rate decay which was tuned over multiple runs. The key insights were that our models did not require many epochs to reach their lowest validation MSE, and that a simple average of the results of two models outperforms each one individually. 

\paragraph{Hyper-parameters of network training}
In each training run of the model, we save only the best model for speed and the best model for steering angle, which are then separately stored and used for predicting values on the test set. 

\paragraph{Model $\textsc{A\_CNN\_single}$} used the Adam optimizer with an initial learning rate of 0.0003. We implemented learning rate decay to 0.0001, 0.00005, and 0.00003 after the first 5, 15, and 20 epochs. Overall the model was trained for 90 epochs, and we used a batch size of 13 for training, validation and testing. The loss criterion used for both speed and steering angle was MSE loss, and the overall model loss was defined as the summation of the speed loss and steering angle loss. 

\paragraph{Model $\textsc{B\_CNN\_stacked}$} was trained in an identical was as the previous model, but the lowest MSE loss was achieved after the 14th epoch. 

\paragraph{Model $\textsc{C\_Bi\_GRU}$} was trained using the Adam optimizer with an initial learning rate of 0.003, which is halved after epochs 20, 30 and 40. We use the same combination loss criterion for as was used in models A and B where we sum the MSE from speed and steering angle. The model was trained for a total of 50 epochs, but the best model was achieved after only the 10th epoch.

The final submission involved taking the average of the results for two of the models used. Doing this caused the MSE for speed to drop from 6.115 to 5.312, and the MSE for steering to drop from 925 to 901, putting us in second place for the competition.

\section{Results}
\label{sec:results}
Our best performing single model for speed was model $\textsc{C\_Bi\_GRU}$ with the lowest MSE on the test set at 6.115, while the best performing single model for steering angle was model Model $\textsc{B\_CNN\_stacked}$ with MSE loss at 925.926. The results are summarized in Table 1. Model B performed the best for steering as a result of the combination of data augmentation which included horizontal flipping of images in a sequence, and the full view of the sequence to the input of the model.

\begin{table}[ht]
\small
\centering
\begin{tabular}{l|c|c}
Model  & MSE Speed & MSE Angle \\ 
\hline
$\textsc{A\_CNN\_single\_image}$ &7.440&\multicolumn{1}{r}{1,140.875}\\ 
$\textsc{B\_CNN\_stacked}$ &7.036&\multicolumn{1}{r}{925.926}\\ 
$\textsc{C\_Bi\_GRU}$  &6.115&\multicolumn{1}{r}{1,075.497}\\ 
$\textbf{Avg:}$ & $\textbf{5.312}$ & \multicolumn{1}{r}{$\textbf{901.802}$}  \\ 
\hline
\end{tabular}
\label{Tab:Res} 
\caption{Results from three models. The best performing model overall is $\textsc{B\_CNN\_stacked}$, which achieved the lowest MSE for steering and was second for speed. The individual model that had the best performance for speed was $\textsc{C\_Bi\_GRU}$. The average of the predictions generated by models B and C provided a significant decrease in the MSE of both speed and steering angle. This combination was our best submission to the competition which awarded us second place in steering angle prediction and second place overall.} 
\end{table}

\section{Conclusions}
\label{sec:conclusions}
In this work, we showed several improvements for predicting steering angle and speed given a sequence of images. First, concatenating segmentation masks with the input images adds side information in the pre-trained segmentation neural network. Second, data augmentation techniques enhanced the performance. Third, stacking the images and segmentation mask sequences into a single input improves the steering predictions significantly. Finally, taking into account multiple models yields our best result. 



\paragraph{Acknowledgements} We would like to thank the Columbia University students of the Fall 2019 Deep Learning class for their participation in the challenge. Specifically, we would like to thank Xiren Zhou, Fei Zheng, Xiaoxi Zhao, Yiyang Zeng, Albert Song, Kevin Wong, and Jiali Sun for their participation.

{\small
\bibliographystyle{ieee_fullname}
\bibliography{bibliography}
}

\end{document}